# Classifications of the Summative Assessment for Revised Bloom's Taxonomy by using Deep Learning

Manjushree D. Laddha[#1], Varsha T. Lokare[*2], Arvind W. Kiwelekar[#3], Laxman D. Netak[#4]

[#] Department of Computer Engineering, Dr. Babasaheb Ambedkar Technological University
Lonere – Raigad (MS), India.

[#] Department of Computer Engineering, Rajarambapu Institute of Technology,
*Rajaramnagar, Islampur – Sangli (MS), India.

[1]mdladdha@dbatu.ac.in, [2]varsha.lokare@ritindia.edu, [3]awk@dbatu.ac.in

***Abstract*** *— Education is the basic step of understanding the truth and the preparation of the intelligence to reflect. Focused on the rational capacity of the human being, the Cognitive process and knowledge dimensions of Revised Bloom's Taxonomy helps to differentiate the procedure of studying into six types of various cognitive processes and four types of knowledge dimensions. These types are synchronized in the increasing level of difficulty. In this paper, Software Engineering courses of B.Tech Computer Engineering and Information Technology offered by various Universities and Educational Institutes have been investigated for Revised Bloom's Taxonomy (RBT). Questions are a very useful constituent. Knowledge, intelligence, and strength of the learners can be tested by applying questions.*

*The fundamental goal of this paper is to create a relative study of the classification of the summative assessment based on Revised Bloom's Taxonomy using the Convolutional Neural Networks (CNN), Long Short-Term Memory (LSTM) of Deep Learning techniques, in an endeavor to attain significant accomplishment and elevated precision levels.*

**Keywords —** *Revised Bloom's Taxonomy, Deep Learning, Software Engineering, Convolutional Neural Networks, Long Short-Term Memory.*

## I. INTRODUCTION

Assessment, of course, is carried out by the tests in which descriptive and multiple-choice questions are present. It is very difficult to find out which particular skills students acquire. If the questions in the test are in agreement with the cognitive process and knowledge dimensions of Revised Bloom's taxonomy, then it is very helpful for teachers to find out which skills students are acquiring and also helpful to set the question paper according to different levels.

Revised Bloom's Taxonomy consists of two dimensions [1] that are cognitive process and knowledge dimensions. Cognitive processes have six different sub-categories. The second dimension describes four knowledge dimensions categories.

Classifying the questions according to these categories manually is a tedious job. Some researchers have classified the questions on cognitive processes, not on knowledge dimensions, by using machine learning classifiers like K-Nearest Neighbor, Logistic Regression, and Support Vector Machine.

In this paper, deep learning techniques are applied to Revised Bloom's Taxonomy, using a dataset of Software Engineering courses. In previous research uses more traditional machine learning algorithms to classify the questions on the cognitive process category. A model is suggested for the classification of question modules established on both cognitive process and knowledge dimensions by using CNN and LSTM.

## II. LITERATURE SURVEY

Section A. presents empirical literature and relevant theoretical regarding RBT to frame the study, along with a layout of how RBT has been used in educational studies.

Section B. presents student's data through the lens of deep learning.

### *A. Revised Bloom's Taxonomy in Education*

RBT is two dimensions in which one dimension is a cognitive process, and another is a knowledge dimension. Both have subcategories again. By using RBT, a framework for conceptual modeling and metamodeling of broad teaching, the correlation between these two dimensions [2] predicts the perception of students such types of problems are solved. In this paper [3], the author proposed a generic framework for conceptual modeling and metamodeling of broad teaching. Based on this framework, a teaching case study is developed, and an evaluation of the teaching case helps to upgrade the teaching of these models in the future. This paper [4] studies the perceptions of students about different orders of thinking skills from aspects of two dimensions, such as Cognitive Process Dimension and Knowledge Dimension of Revised Bloom's Taxonomy.

The classification of questions into these two dimensions, which have subcategories, is done manually [5]. By applying the machine learning methods SVM and K-NN, the classification of questions into the cognitive process of RBT is carried-out. This helps teachers to set the question paper according to different cognitive levels. For students, it helps to enhance their ability and thinking and to solve the questions. The feature is extracted by using the active verbs from the questions [6]. In some papers, the features selection is carried out by TFPOS-IDF and word2vec [7] and by using Logistic Regression, K-Nearest Neighbour, and Support Vector Machine to classify the questions based on cognitive process.

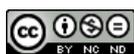

### *B. Deep Learning in Education*

Deep learning techniques are used for processing [8] video, images, audio, and speech, whereas it has shown a light on sequential data such as text also. A detailed study [9] of DL techniques is discussed in this work, beginning with an introduction, the kind of architectures applied in each function, an analysis of the hyperparameter design, and a catalog of the current structure to assist in the implementation of DL models. Since discussing DL architecture depends on factual processes, the information provided can suggest future developments of DL applications.

By applying deep learning without the use of sensors [10], to detect students' affective states, student information is utilized. For detecting affective states of students, recurrent neural network applications are used on the data collected from the online learning platform.

In Education, Data Mining and Deep learning importance are highlighted in finding academically weak students. By helping them for their performance improvement by providing various recommendation systems [11]. The quick prediction of students at-risk of poor performance [12] by using clicks stream data, and discovers important features that improve student to exceed others. The deep learning model predicts the students' performance. Based on this, it suggests intervention. The university executes corrective measures for student counseling and support. Deep learning helps the learners with their upcoming ideas by recognizing the major unseen design from their previous information of assessments [13].

Not only with the assessment of multiple-choice questions, by using the programming exercise with the help of drag and drop, but a block of code is also used to predict the future student performance by using DL [14].

The common methods used in text feature extraction like filtration, fusion, mapping, and clustering method. It then enlarges normally used deep learning techniques in text feature extraction methods [15] like Convolution Neural Network (CNN) and Recurrent neural network (RNN). Deep learning models that have excelled machine learning techniques in numerous text classification methods.

The text classification tasks are question answering, and sentimental analysis, topic analysis, news categorization. Text Classification (TC) is the procedure of sorting texts. These classifications are carried out by DL-based text classifiers [16]. Not only the text classification but it also classifies the text document with the sequences of words into different categories such as courses of programs, different religions, and laws, by using LSTM (long short term memory) [17]. LSTM uses in duplicate identification of questions and text [18].

In this paper, by using deep learning methods like CNN and LSTM, classify the summative assessment of the software engineering course as per the Revised Bloom's Taxonomy. A model is proposed in this research work for the classification of question items based on both cognitive process and knowledge dimensions by using CNN and LSTM.

### III. PROPOSED MODEL OF CLASSIFICATION OF SUMMATIVE ASSESSMENT FOR REVISED BLOOM'S TAXONOMY BY USING CNN AND LSTM

The summative assessment as per revised bloom's taxonomy paradigms has been considered as a multiclass classification problem. As per revised Bloom's Taxonomy, Cognitive processes and a knowledge dimension play an important part in the analysis of the assessment strategies. As shown in Figure 1, the suggested model can predict the cognitive process along with the knowledge dimension of the given question. Cognitive processes are further categorized into a total of six parts, namely Remember, Understand, Apply, Evaluate, Analyze, and Create. Also, the Knowledge dimension is further subdivided into a total of four dimensions: Factual, Conceptual, Procedural, and Meta - Cognitive. Here, in this research, the fourth dimension Meta Cognitive has not been considered as it depends on the individual perspective.

Here, a total of 844 questions from the Software Engineering course have been considered for analysis purposes. The supervised learning approach is considered here, along with the labeled cognitive processes and knowledge dimension for every input question. The Pre-processing is done using NLTK libraries to remove punctuations, special symbols, etc. The Natural Language Processing (NLP) technique is used for data cleaning purposes. Finally, the pre-processed data has been supplied as an input to the deep learning models, namely CNN and LSTM. A total of 70% of data (591 Questions) has been used for training, and the rest 30% (253 Questions) data has been considered for testing purposes. Once the training process gets completed, the CNN and LSTM based model is ready to predict the cognitive process and knowledge dimension for the new questions.

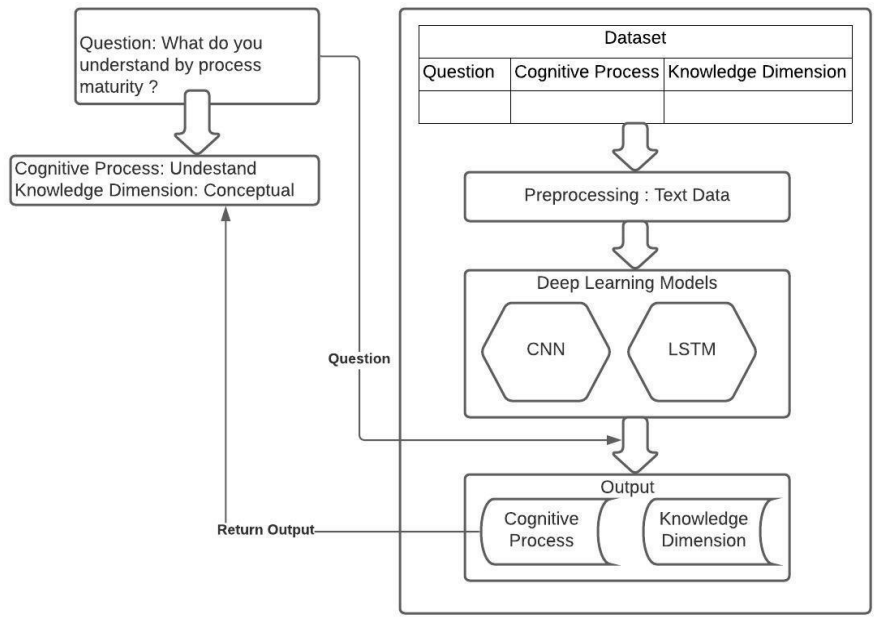

**Figure 1: Model of summative assessment for Revised Bloom's Taxonomy by using CNN and LSTM**

*A. Revised Bloom's Taxonomy*

Revised Bloom's Taxonomy consists of two-dimension matrix columns representing the cognitive process, and rows represent knowledge dimensions. Cognitive processes have six different sub-categories remember, understand, apply, analyze, evaluate, and create. The second dimension specifies four knowledge dimensions categories that are Factual, Conceptual, Procedural, and Meta-cognitive. Knowledge dimensions are the assimilation of information through learning, while a cognitive skill means the capacity to relate knowledge and use that knowledge to solve problems and write descriptive answers. Cognitive skills are acquired by training.

*i) Cognitive Processes:*

Generally, in the questions, the verb represents the cognitive processes. The cognitive process is categorized into six categories. While answering the question, if a student has to recall a particular concept, then that question comes under the remember category. In this way, the questions are categorized according to the learning activity.
a) Remember - recognizing (identifying), recalling (retrieving).
b) Understand - classifying (categorizing, subsuming), comparing (contrasting, mapping, matching), explaining (constructing models).
c) Apply – executing (carrying out), implementing (using).
d) Analyze – differentiating (discriminating, distinguishing, focusing, electing), organizing (finding, coherence, integrating, outlining, parsing, structuring).
e) Evaluate - checking (coordinating, detecting, monitoring, testing), critiquing (judging).
f) Create - generating (hypothesizing), planning (designing), producing (construct)

*ii) Knowledge Dimensions:*

The noun normally represents knowledge. The knowledge dimensions are of four types that are factual, conceptual, procedural, and meta-cognitive. The meta-cognitive is not considered, as it is the ability of one's own (individual) thinking.
1) Factual Knowledge Dimension:
   The basic knowledge required for the software engineering course. Basic knowledge is like terminology, specific details, and lists out. From the already learned information to extract the exact answer to the question. Consider a question from the dataset is:
   Explain the role of product, process, and people in project management.
   From the students, it is expected that to extract information which they already know from the concept. Elaborate something from the knowledge of that concept is required. It should not be explained from scratch. Knowledge of specific details is required to perceive the meaning of content. The content may be text, graphics, or voice. To demonstrate the meaning by interpreting, explaining, classifying, and summarizing ideas and concepts. So it is under the category understand-factual.
2) Conceptual Knowledge Dimension:
   From the learned information, try to compare concepts with other concepts, classifications, categories, principles and generalizations, models, theories, and structures. Consider a question from the dataset is:

Compare conventional technique and object-oriented technique to software development?

From the students, it is expected that they should understand those concepts in detail. Interpret the conventional approach and object-oriented approach precisely so that students can understand what that concept wants to say. Breaking down the concepts into minute parts helps the students to take useful insights from them and to find out how they are related to one another. So it is under the category analysis-conceptual.

3) Procedural Knowledge Dimension

Specific knowledge of the subject is essential so that step by step procedure can explain techniques, algorithms, methods, and criteria for using particular skills. . Consider a question from the dataset is:

List down and explain the activities of scheduling and tracking for the Library management system.

From the students, it is expected that they should know these two concepts completely. Interpret the conventional approach and object-oriented approach precisely so that students can understand what that concept wants to say. Breaking down the concepts into minute parts helps the students to take useful insights from them and to find out how they are related to one another. So it is under the category analysis-conceptual.

From the students, it is expected that they should be familiar with the generalized activities of scheduling and tracking. Then they can implement these activities on the Library management system. Applying has two subcategories in RBT, performing and accomplishing. Performing means using established knowledge in operation (so is an intimate function), whereas accomplishing means using it in an issue (which inclined to be an unfamiliar function).

*B. Deep Learning Models*

Here a total of two deep learning models are used for experimental and analysis purposes. CNN and LSTM models are used for many applications, including text data analysis [19, 20], for the classification of the question these models have considered. The description of each model is given below:

i) *Convolutional Neural Network model (CNN):*

This CNN model has been used in many applications ranging from image data to text data. Even though this model is designed for image classification, it shows prominent results in text classification problems also. Convolution networks automatically identify local indicators in the question that can predict the correct cognitive and knowledge level. Here, the position of the indicator is not important. For example, consider the following scenario:

Question 1: What is the work breakdown structure? How is this used in the task network of project scheduling?

Question 2: How do you measure the quality of software? Explain with any two quality metrics. The cognitive level of the above two questions is Apply. CNN models identify local indicators like how, used and measured, etc., to identify the cognitive level of the question. Similarly, the knowledge dimensions of the questions are predicted.

Input: Question
Apply following steps in preprocessing:

a) **Tokenizer:** Splits sentences into words. It returns the frequently appeared words along with their frequency and removes the less frequently appeared words. It also converts lowercase words to the uppercase word.
Below is the code snippet for the same.
*tokenizer = Tokenizer(num_words = n_most_common_words, filters = '!"#$%&()*+,-./:;<=>?@[\]^_`{|}~', lower = True)*
*tokenizer.fit_on_texts(concated['Question'].values)*

b) **Pad Sequence:** The deep learning model is working on the principle of the same common length of the training text input. Hence, max length is considered here for each input with the same length.
*X = pad_sequences(sequences, maxlen = max_len)*

c) **Label Encoding the Target Variable:**
Here, the input is in categorical form, as questions with different cognitive levels are considered as an input to the model. As deep learning models require input in numerical form, there is a need for conversion of this categorical data to numerical data. There are two methods *to* change categorical data to numerical data, namely integer encoding and one hot encoding. If no ordinal relationship exists between the texts, integer encoding will not be that much useful.

Here, one hot encoding technique is applied as there is a relationship that exists between the tokens present in the text input.

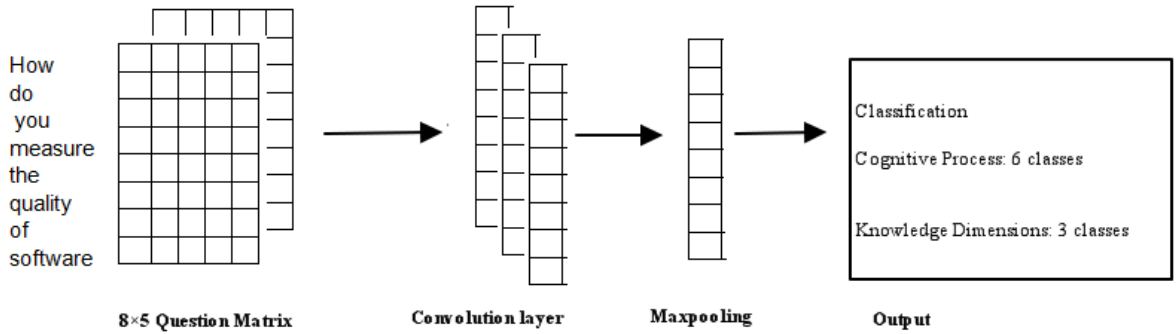

**Figure 2: CNN Architecture**

*One-hot encode the label*
*concated.loc[concated['Cognitive_Category']=='Remember', 'LABEL'] = 0*
*concated.loc[concated['Cognitive_Category']=='Understand', 'LABEL'] = 1*
*concated.loc[concated['Cognitive_Category']=='Apply', 'LABEL'] = 2*
*concated.loc[concated['Cognitive_Category']=='Evaluate', 'LABEL'] = 3*
*concated.loc[concated['Cognitive_Category']=='Analyze', 'LABEL'] = 4*
*concated.loc[concated['Cognitive_Category']=='Create', 'LABEL'] = 5*

[1. 0. 0. 0. 0. 0.] Remember
[0. 1. 0. 0. 0. 0.] Understand
[0. 0. 1. 0. 0. 0.] Apply
[0. 0. 0. 1. 0. 0.] Evaluate
[0. 0. 0. 0. 1. 0.] Analyze
[0. 0. 0. 0. 0. 1.] Create

Similarly for Knowledge Dimension:
#One-hot encode the lab
*concated.loc[concated['Knowledge_Dimension'] == 'Factual', 'LABEL'] = 0*
*concated.loc[concated['Knowledge_Dimension'] == 'Conceptual', 'LABEL'] = 1*
*concated.loc[concated['Knowledge_Dimension'] == 'Procedural', 'LABEL'] = 2*
[1. 0. 0. ] Factual
[0. 1. 0. ] Conceptual
[0. 0. 1. ] Procedural

**d) Simple CNN architecture to classify questions into cognitive processes and knowledge:** As shown in Figure 2, the input is first preprocessed and supplied to the CNN model. Here the sample question considered is: "How do you measure the quality of the software." Total 5 dimensions are considered, and the matrix of 8 by 5 questions has been input for the CNN model. The imputed matrix then passes through the convolution layer, in which filtering of the features is done. Max Pooling function is applied to the selected features, and finally, the sigmoid function is applied for the correct prediction of the cognitive process and knowledge dimension.

ii) **Long Short Term Memory (LSTM) Model:**
As shown in Figure 3, the LSTM model is used for the prediction of multiclass text classification problems. Here, the preprocessed question vectors are passed through each LSTM layer with dimension = 4. The first Embedding layer contains arguments like several common words, embedding dimensions, etc. This is needed to represent each word in the question as a vector. Next, the SpatialDropout1D is used for dropouts with 0.7 values for processing NLP data. LSTM layer is used with 10 memory units, and finally, the softmax activation function is used for the prediction of the cognitive process and knowledge dimension of the input question. As this is the multiclass classification problem, for the loss function, categorical cross-entropy is used.

*model.add(Embedding(n_most_common_words, emb_dim, input_length=X.shape[1]))*
*model.add(SpatialDropout1D(0.7))*
*model.add(LSTM(10, dropout = 0.7, recurrent_dropout = 0.7))*
*model.add(Dense(3, activation='softmax'))*

### IV. EXPERIMENT AND DISCUSSION
To increase the accuracy of student learning assessment, the Revised Bloom's Taxonomy has
been introduced [21]. Unlike the older taxonomy, the newer one is two-dimensional and mainly focused on six cognitive processes and 4 knowledge dimensions, as shown in Table I.

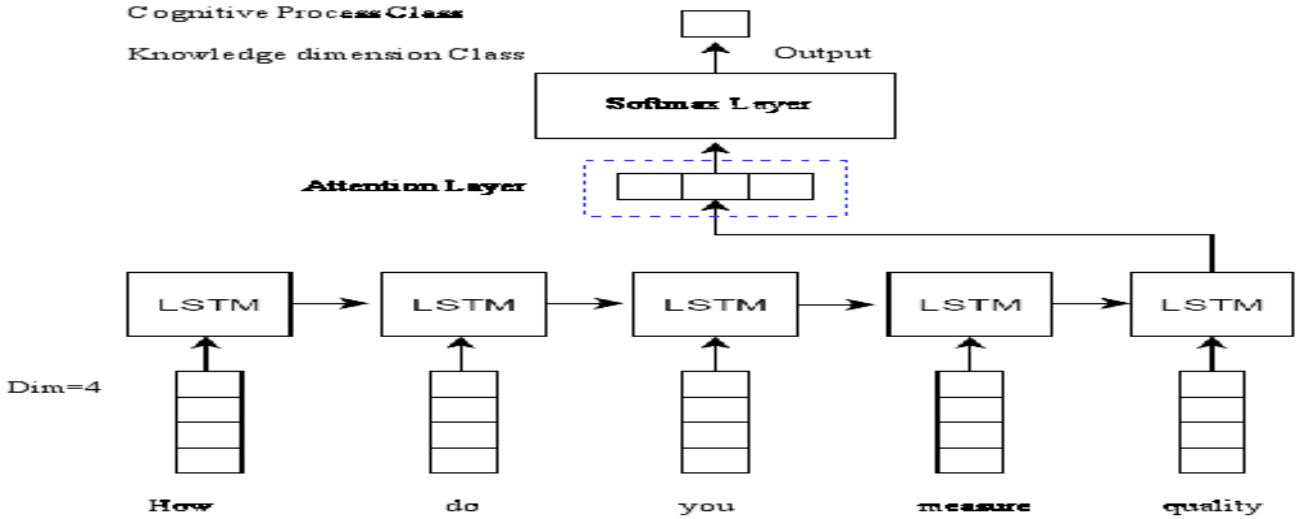

**Figure 3: LSTM Architecture**

*A. Dataset:*

Total 844 questions for the subject Software Engineering have been considered as an input along with Revised Bloom's Taxonomy. The questions are collected from the question papers of various academic institutions. In some institutions, the Software Engineering course is provided in the fourth or fifth semester, as shown in Table I.

*B. Result Analysis:*

As shown in Figure 4-7 and Table II, it is observed that the CNN model gives better accuracy in the classification of cognitive processes as compared to the LSTM model. Also, the loss is comparatively less in the CNN classifier. As shown in Table II, the Training and Testing accuracy of the CNN model is 75% and 80%, respectively. As the number of epochs rises, precision in the training and testing phase also increases. The loss is increased in the Testing phase as compared to the Training phase for both models.

The accuracy parameter is more in the testing phase (80%) for CNN than the training phase (75%), while in the case of the LSTM model, the training phase (77%) has shown better accuracy than the testing phase (71%). In the case of knowledge dimension prediction CNN. The classifier has shown better performance than LSTM in the testing phase, but in the training phase, LSTM has shown 94.44% accuracy while CNN has only 88.89% accuracy. In case of loss, CNN has shown better results than LSTM.

**Table I: Dataset for Cognitive Processes and Knowledge Dimensions**

| Dataset | Cognitive Processes | | | | | | Knowledge Dimensions | | |
|---|---|---|---|---|---|---|---|---|---|
| | Remember | Understand | Apply | Evaluate | Analyze | Create | Factual | Conceptual | Procedural |
| No. of Questions | 202 | 464 | 61 | 29 | 41 | 47 | 320 | 290 | 234 |

*C. Measuring Parameters:*

To compute the performance of the model, a total of two computing parameters has been examined, namely Accuracy and Loss.

 a) Accuracy: Accuracy is based on the ratio of correctly predicted class labels and the total number of the testing labels.

 b) Loss: As this proposed model is based on the concept of a multiclass classification problem, the cross-entropy loss function is applied to measure the overall loss.

$$CE = -\sum_{i}^{c} t_i log(s_i) \quad \text{---------- (1)}$$

Where $t_i$ and $s_i$ are the ground truth and the CNN score for each classification for the total number of *C* classes.

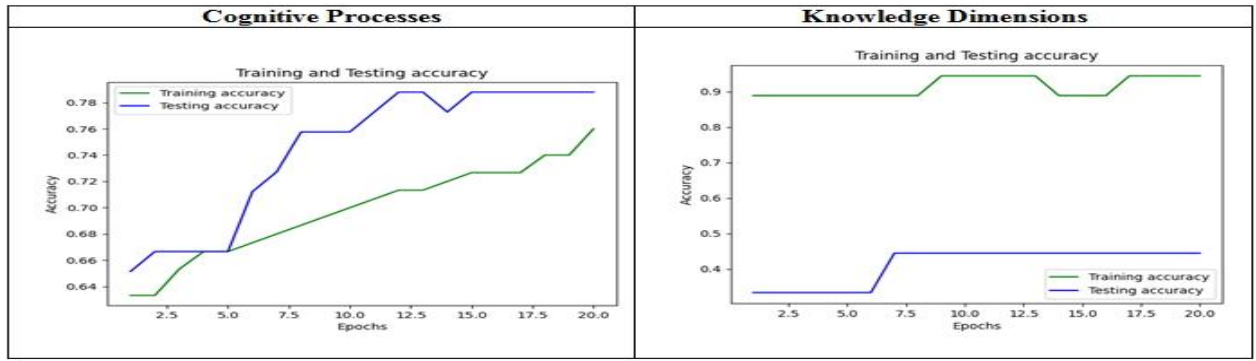

**Figure 4: Accuracy of CNN model**

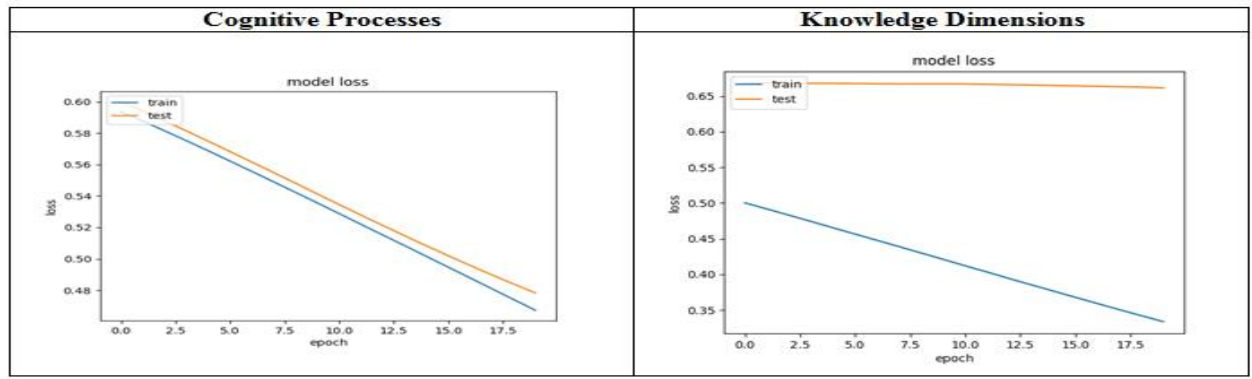

**Figure 5: Loss of CNN model**

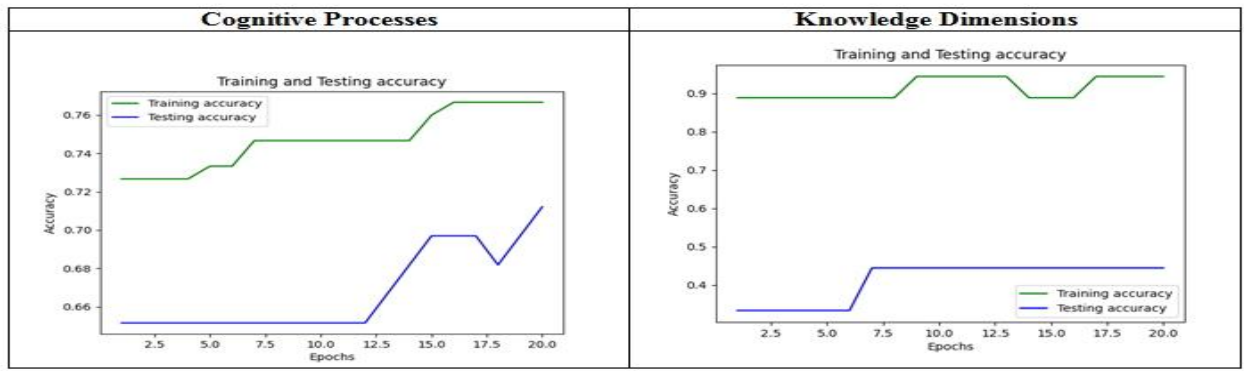

**Figure 6: Accuracy of LSTM model**

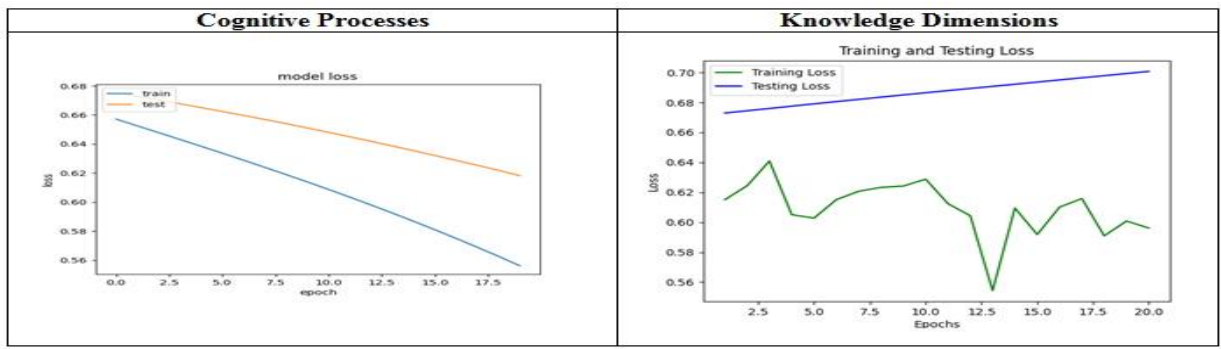

**Figure 7: Loss of LSTM model**

**TABLE II: Comparative Results**

| Model | Cognitive Process | | | | Knowledge Dimension | | | |
|---|---|---|---|---|---|---|---|---|
| | Accuracy | | Loss | | Accuracy | | Loss | |
| | Training | Testing | Training | Testing | Training | Testing | Training | Testing |
| CNN | 75% | 80% | 0.46 | 0.47 | 88.89% | 66.67% | 0.33 | 0.66 |
| LSTM | 77% | 71% | 0.57 | 0.63 | 94.44% | 44% | 0.55 | 0.70 |

## CONCLUSION

A deep learning-based model is suggested in this paper for the prediction of cognitive processes and knowledge dimensions. Here, for the analysis intention, the questionnaire from the software Engineering course has been considered for the experimental work. It has been observed that the proposed CNN model has given 80% accuracy in the prediction of cognitive processes among the categories like Understand, Remember, Apply, etc. In the case of knowledge dimensions prediction, LSTM is better in the training phase than the CNN model. But for the testing phase, CNN has been shown 66.67% accuracy, and LSTM has given only 44% accuracy. In case of loss also CNN has shown better performance than LSTM. Hence, the results obtained by the LSTM models are comparatively not that promising.

It helps teachers to find out in which particular cognitive process and knowledge dimension students are lagging. For students, it helps to cope up with the problematic skill, and in the Universities, it suggests making a policy decision to set the question paper according to different levels like easy, medium, hard based on classification of the questions.

In the future, additional courses from different programs can be examined for generalizing the classification connecting the knowledge dimension and cognitive process. This work can be further extended by considering the remaining knowledge dimension category, i.e., the Meta-cognitive domain.